# A Nonparametric Adaptive EWMA Control Chart for Binary Monitoring of Multiple Stream Processes

Faruk Muritala[1a], Austin Brown[1b], Dhrubajyoti Ghosh[1c], Sherry Ni[1d]
[1]School of Data Science and Analytics, Kennesaw State University, Kennesaw, United States of America.

**Corresponding Author**
[a]Corresponding author email: [fmurital@students.kennesaw.edu]

**ORCIDs**
[a]ORCID: 0000-0002-5857-9874
[b]ORCID: 0000-0003-1530-0009
[c]ORCID: 0000-0002-3360-3786
[d]ORCID: 0000-0003-0634-0025

**ABSTRACT**

Monitoring binomial proportions across multiple independent streams is a critical challenge in Statistical Process Control (SPC), with applications from manufacturing to cybersecurity. While EWMA charts offer sensitivity to small shifts, existing implementations rely on asymptotic variance approximations that fail during early-phase monitoring. We introduce a Cumulative Standardized Binomial EWMA (CSB-EWMA) chart that overcomes this limitation by deriving the exact time-varying variance of the EWMA statistic for binary multiple-stream data, enabling adaptive control limits that ensure statistical rigor from the first sample. Through extensive simulations, we identify optimal smoothing ($\lambda$) and limit (L) parameters to achieve target in-control average run length ($ARL_0$) of 370 and 500. The CSB-EWMA chart demonstrates rapid shift detection across both $ARL_0$ targets, with out-of-control average run length ($ARL_1$) dropping to 3–7 samples for moderate shifts ($\delta=0.2$), and exhibits exceptional robustness across different data distributions, with low $ARL_1$ Coefficients of Variation (CV < 0.10 for small shifts) for both $ARL_0$ = 370 and 500. This work provides practitioners with a distribution-free, sensitive, and theoretically sound tool for early change detection in binomial multiple-stream processes.

**Keywords:** Statistical Process Control, EWMA, multiple-stream processes.

## Introduction

Statistical Process Control (SPC) remains a cornerstone methodology for monitoring and improving quality across diverse sectors, including manufacturing, healthcare, cybersecurity, and service industries (Montgomery 2019). Since inception by Shewhart in the 1920s, the core objective of SPC has been to distinguish between common cause variation inherent to a stable process and special cause variation that signals a process change requiring intervention (Shewhart, 1931). Traditional control charts form the primary toolkit for this purpose, with Shewhart charts (Shewhart, 1931), Cumulative Sum (CUSUM) charts (Page, 1954), and Exponentially Weighted Moving Average (EWMA) charts (Lucas and Saccucci, 1990) representing foundational approaches. Comprehensive reviews of these methodologies highlight their evolution and widespread application in the area of modern quality and reliability monitoring, including high-volume manufacturing, healthcare surveillance, cyber-intrusion detection, and service-system performance assessment (Montgomery 2019; Woodall 2000).

However, these classical methods are built upon foundational assumptions of normality and independence in process data, assumptions frequently violated in real-world applications involving discrete outcomes, such as the monitoring of binomial proportions (e.g., defect rates, infection rates, or fraud indicators). In addition, current literature reveals a substantial gap in exact statistical formulations for EWMA charts applied to binomial processes. The commonly used asymptotic variance, $Var = \sigma^2 \sqrt{\frac{\lambda}{2-\lambda}}$, fails to capture the time-varying nature of the variance during initial monitoring phases (Anastasopoulou and Rakitzis 2022). This limitation becomes particularly problematic in applications where early detection of process changes is critical, such as in manufacturing quality control, cybersecurity threat detection, or healthcare outcome monitoring (Best and Neuhauser 2006; Schmidtke et al. 2024).

For attribute data, the p-chart and np-chart are the standard Shewhart tools. While simple to implement, they are notably insensitive to small, sustained shifts, and their performance is heavily dependent on sample size. More critically, they are univariate in nature and do not scale directly to Multiple Stream Processes (MSPs), a common scenario in modern industry where several parallel lines, machines, or locations must be monitored simultaneously (Boyd 1950).

The challenge of MSP monitoring originated from practical needs in multi-head filling machines and multi-spindle operations, where traditional single-stream charts proved inadequate (Boyd 1950; Abdulaziz, et al. 2019). However, early MSP approaches, such as those based on range statistics, often relied on normality assumptions that limited their robustness (Mortell and Runger 1995). Recognizing these limitations, Runger, Alt, and Montgomery (1996) employed principal components analysis to decompose MSP measurements into orthogonal components, enabling separate monitoring of overall process shifts and individual stream deviations. This two-component variance model proved more effective than range-based statistics for detecting stream-specific changes. Building on this foundation, researchers explored alternative monitoring strategies. Meneces et al. (2008) revisited the use of individual Shewhart charts for each stream, demonstrating that with appropriate Bonferroni-adjusted control limits, this approach effectively utilizes stream-specific information. Liu, MacKay, and Steiner (2008) proposed F-test and likelihood ratio-based methods, while Epprecht, Barbosa, and Simões (2011) introduced a modified group chart based on deviations from the grand mean, showing enhanced detection for moderate shifts. A comprehensive review of MSP control techniques from 1950 through 2015 is provided by Epprecht (2015), which identifies key developments and remaining research gaps in the field.

A pivotal shift toward distribution-free MSP monitoring occurred with the introduction of nonparametric methods. Unlike parametric methods that require normality assumptions and can suffer from inflated Type I error rates when these assumptions are violated, nonparametric approaches maintain their statistical validity across diverse distributions, providing robust performance without requiring knowledge of the underlying process distribution (Chakraborti and Graham 2019). The theoretical foundations for using sign-based statistics in process control are well-established (Chakraborti and Graham 2019; Chakraborti, van der Laan, and Bakir 2001). When observations are dichotomized relative to an in-control median, the resulting binary indicators follow a known binomial distribution regardless of the underlying process distribution, enabling distribution-free control limits with exact false alarm rates. To the best of our knowledge, the work by Brown and Schaffer (2021) stands as the primary contribution to nonparametric control charting for MSPs. Their Nonparametric Extended Median Test CUSUM (NEMT-CUSUM) chart successfully circumvents distributional assumptions by dichotomizing stream data relative to the in-control median, thereby reducing the MSP monitoring problem to one of tracking a binomial proportion.

Despite its innovation, this pioneering approach focuses primarily on detecting larger shifts and relies on asymptotic properties requiring substantial data, making it less suitable for detecting subtle, early changes. Additionally, their approach would not be appropriate in small sample or individual observation situations, which are common in modern processes where all units are measured. Recent empirical work has further demonstrated the importance of real-time, small-sample monitoring approaches, including the successful application of SPC-based CUSUM methods to detect early performance deterioration and reinjury risk among Major League Baseball pitchers (Brown, Muritala and Haller 2026). Meanwhile, the EWMA framework offers particular advantages in multiple stream contexts through its ability to accumulate information over time while maintaining computational tractability (Lucas and Saccucci 1990). The standard EWMA recursion provides adjustable memory through the smoothing parameter $\lambda$, enabling customization of detection sensitivity. For multiple stream applications, this memory-based approach proves especially valuable in detecting gradual degradation.

However, the integration of EWMA charts into the nonparametric MSP framework remains largely unexplored. Existing EWMA applications for attributes in single-stream contexts (Anastasopoulou and Rakitzis 2022; Gan 1993) rely on the asymptotic variance approximation, which fails to provide accurate control limits during the initial phase of monitoring when data is limited and the variance of the EWMA statistic is still evolving (Lucas and Saccucci 1990; Anastasopoulou and Rakitzis 2022). This creates a dual gap: first, in the application of the more sensitive EWMA framework to the nonparametric MSP problem, and second, in the derivation of exact, time-varying control limits for this application to ensure optimal performance from the very first observation.

This paper makes three primary contributions to address these gaps. First, we introduce the Cumulative Standardized Binomial EWMA (CSB-EWMA) chart, a novel distribution-free tool for monitoring Multiple Stream Processes. Second, we derive the exact, time-varying variance of the EWMA statistic for binary data, enabling adaptive control limits that are statistically rigorous from the first sample and still hold even in the case of individual observations. Third, through comprehensive simulation, we identify optimal chart parameters and demonstrate the method's superior sensitivity and robustness compared to existing approaches.

## Methodology

This section introduces the structural and theoretical basis of the Cumulative Standardized Binomial EWMA (CSB-EWMA) control chart. Before exploring its mechanics, it is important to clarify the assumptions that support its application. The charting method presumes that the $k$ data streams exhibit temporal independence. Within any single stream, observations collected at a given time point are also assumed to be mutually independent. Additionally, the process under surveillance must be measurable on at least an ordinal scale (Brown and Schaffer 2021).

Let $y_{ij}$ denote the $j$th observation randomly sampled from the stream $i$ which has an unknown probability distribution but a known, in-control median, denoted by $\tilde{\mu}_0$. Here, $i = 1, 2, \ldots k$, and $j = 1, 2, \ldots t$. Now, a binary indicator variable is then defined as $x_{ij} = I(y_{ij} \geq \tilde{\mu}_0)$, where $x_{ij} = 1$ if the inequality is true, and 0 otherwise. Provided that the median of stream $i$ remains stable at $\tilde{\mu}_0$, the indicator variable $x_{ij}$ follows a binomial distribution with parameters $n = 1$ and $p_0 = 0.5$, as shown below:

$$x_{ij} \sim BIN(n = 1, p_0 = 0.50) \tag{1}$$

Both the raw observations $y_{ij}$ and their corresponding binary indicators $x_{ij}$ are collected across all $k$ streams. Note, because it is assumed that for a given time point $j$, each of the observations between the streams are mutually independent Bernoulli random variables, then it is straightforward to see that:

$$C_j = \sum_{i=1}^{k} x_{ij} \sim BIN(n = k, p_0 = 0.50) \tag{2}$$

The cumulative count of exceedances up to time $t$ is:

$$Q_t = \sum_{j=1}^{t} \sum_{i=1}^{k} x_{ij} \sim BIN(kt, p_0) \tag{3}$$

The standardized statistic is:

$$W_t = \frac{Q_t - kp_0 t}{\sqrt{kp_0(1-p_0)t}} \tag{4}$$

The EWMA statistic at time $t$ is as follows:

$$r_t = \lambda W_t + (1-\lambda) r_{t-1}, \quad r_0 = 0, \quad 0 < \lambda \leq 1 \tag{5}$$

Here $\lambda$ is the EWMA smoothing parameter, selected through simulation or historical in-control data. Then, we can derive the time-varying $r_t$ as:

$$E[r_t] = (1-\lambda)^t r_0,$$
$$Var(r_t) = \lambda^2 \left[ 2 \sum_{j=1}^{t-1} \sum_{i=j+1}^{t} (1-\lambda)^{2t-i-j} * \frac{\sqrt{j}}{\sqrt{i}} + \sum_{i=1}^{t} (1-\lambda)^{2t-2i} \right] \tag{6}$$

and the time-varying control limits are given as:

$$UCL_t = (1-\lambda)^t r_0 + L\sqrt{Var(r_t)}, \quad LCL_t = (1-\lambda)^t r_0 - L\sqrt{Var(r_t)} \tag{7}$$

where $L$ is the control-limit multiplier. As $t \to \infty$, $Var(r_t) \to 1$, yielding steady-state limits $\pm L$.

Detailed derivations of expressions (6) and (7) (see also Muritala et al. 2026 for the complete mathematical framework). These derivations were validated via Monte Carlo simulation, confirming the accuracy of the theoretical mean and variance expressions under in-control conditions.

## Optimal Parameter Determination and ARL Performance Analysis

The Average Run Length (ARL) is the primary and traditional metric for assessing control chart performance. Specifically, $ARL_0$ denotes the expected number of samples until a false alarm when the process is in control, and $ARL_1$ represents the expected number of samples required to detect a sustained process shift of magnitude δ, where $δ = p_1 - p_0$ indicates the change in the monitored binary proportion.

The optimization objective is to minimize $ARL_1$ for a range of shift sizes while maintaining a predefined $ARL_0$ target. Following industry standards, $ARL_0$ values of 370 (analogous to traditional 3-sigma limits) and 500 (representing stricter false alarm control) are used (Montgomery 2019; Lucas and Saccucci 1990). The optimization framework involves a comprehensive grid search across the parameter space to identify combinations that achieve the target $ARL_0$, followed by an evaluation of their out-of-control detection performance and a validation of their robustness across different data distributions.

### 3.1 Simulation: Experimental Design for Parameter Optimization

A full factorial simulation design was implemented across five critical dimensions to ensure an exhaustive exploration. The core parameters are summarized in Table 1.

Table 1: Simulation Design Parameters for Optimization

| Parameter | Values |
|---|---|
| Smoothing Constant (λ) | 0.10 to 0.90 (step 0.025) |
| Control Limit Multiplier (L) | 1.00 to 2.50 (step 0.025) |
| Number of Streams (k) | 2, 3, 5, 10, 15, 20, 50, 100 |
| Shift Magnitude (δ) | 0.05 to 0.50 (step 0.05) |
| In-Control Proportion ($p_0$) | 0.50 |
| Monte Carlo Replications | 10,000 ($ARL_0$); 50,000 ($ARL_1$) |

### 3.2 Optimal In-Control Parameter Selection

An extensive grid search was conducted to identify (λ, L) pairs yielding $ARL_0$ values approximating the targets of 370 and 500. Representative optimal combinations are presented in Table 2.

Table 2: Optimal (λ, L) Combinations for Target $ARL_0$ Values

| λ Range | Target $ARL_0$ | Optimal λ | Optimal L | Achieved $ARL_0$ | Target $ARL_0$ | Optimal λ | Optimal L | Achieved $ARL_0$ |
|---|---|---|---|---|---|---|---|---|
| 0.1–0.2 | 370 | 0.175 | 1.375 | 372.32 | 500 | 0.150 | 1.525 | 501.31 |
| 0.2–0.3 | | 0.200 | 1.400 | 375.21 | | 0.200 | 1.575 | 500.80 |
| 0.3–0.4 | | 0.350 | 1.500 | 373.90 | | 0.300 | 1.650 | 501.11 |
| 0.4–0.5 | | 0.400 | 1.525 | 374.48 | | 0.450 | 1.725 | 500.69 |
| 0.5–0.6 | | 0.550 | 1.575 | 368.83 | | 0.525 | 1.750 | 501.07 |
| 0.6–0.7 | | 0.650 | 1.625 | 373.83 | | 0.650 | 1.800 | 505.32 |
| 0.7–0.8 | | 0.725 | 1.650 | 372.99 | | 0.700 | 1.800 | 502.93 |
| 0.8–0.9 | | 0.800 | 1.675 | 367.53 | | 0.850 | 1.875 | 501.88 |
| 0.9–1.0 | | 0.900 | 1.700 | 360.11 | | 0.925 | 1.875 | 501.88 |

The parameter relationships reveal important patterns. A direct positive λ-L relationship was observed, where smaller λ values (implying greater memory and smoother statistics) require correspondingly smaller L values to maintain the same $ARL_0$. This relationship progresses in a near-linear fashion across the range of λ.

### 3.3 $ARL_1$ Methodology for Shift Detection

The detection performance was evaluated across shift magnitudes $δ = 0.05$ to $0.50$ in increments of 0.05,

where $\delta$ represents the deviation from the in-control proportion $p_0 = 0.50$ to $p_1 = p_0 + \delta$. For each optimal $(\lambda, L)$ combination identified in Table 2. The $ARL_1$ calculations followed the same dichotomization approach described in the derivation, where continuous data from across four distinct distributions: Normal, Laplace (symmetric heavy-tailed), Uniform (bounded), and Exponential (skewed) were thresholded using the in-control median.

### 3.4 $ARL_1$ Performance Tables by Distribution

The detection capability of the optimal parameter sets was evaluated by computing $ARL_1$ for shifts from $\delta = 0.05$ to $0.50$. Figure 1 illustrates the performance across the different distributions under the $ARL_0$ targets of 370 and 500, respectively

Figure 1 (a)

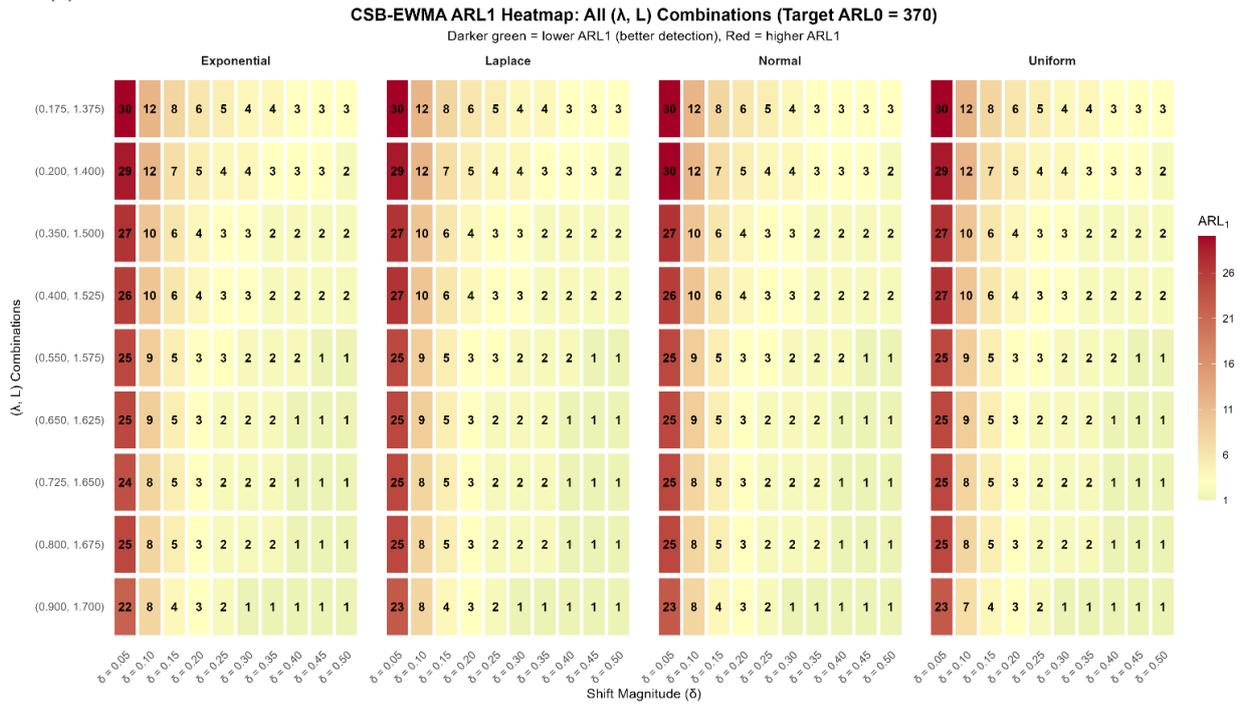

Figure 1 (b)

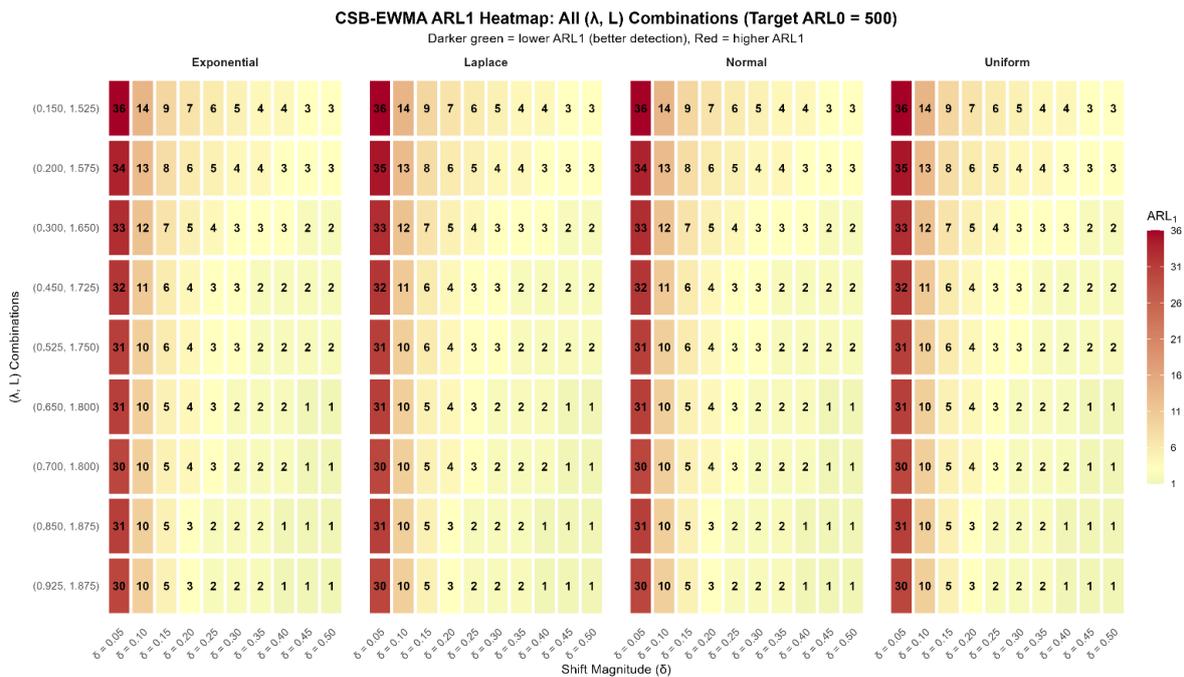

*Figure 1 (a) & (b): ARL$_1$ Values across distributions for the ARL$_0$ targets of (a) 370 and (b) 500

The results demonstrate a strong inverse monotonic relationship where ARL$_1$ decreases consistently as shift magnitude δ increases across the distributions. For small shifts (δ = 0.05), detection requires 23–30 and 30–36 samples on average for ARL$_0$ targets of 370 and 500, respectively, which represents a marked improvement over the traditional memoryless p-chart (Montgomery 2019). For ARL$_0$ targets of 370, moderate shifts (δ = 0.20), ARL$_1$ drops to 3–6, enabling timely intervention, and for large shifts (δ ≥ 0.30), most combinations achieve ARL$_1$ ≤ 3, with several reaching the theoretical minimum of 1, indicating near-immediate detection capability.

## Results

*4.1: Coefficient of Variation (CV) of ARL$_1$ Across Shift Magnitudes*

To further quantify the consistency of performance across the entire range of shift magnitudes, the Average Coefficient of Variation (CV) of ARL$_1$ across the four distributions was calculated. The CV provides a normalized measure of variability, where a lower value indicates more consistent performance.

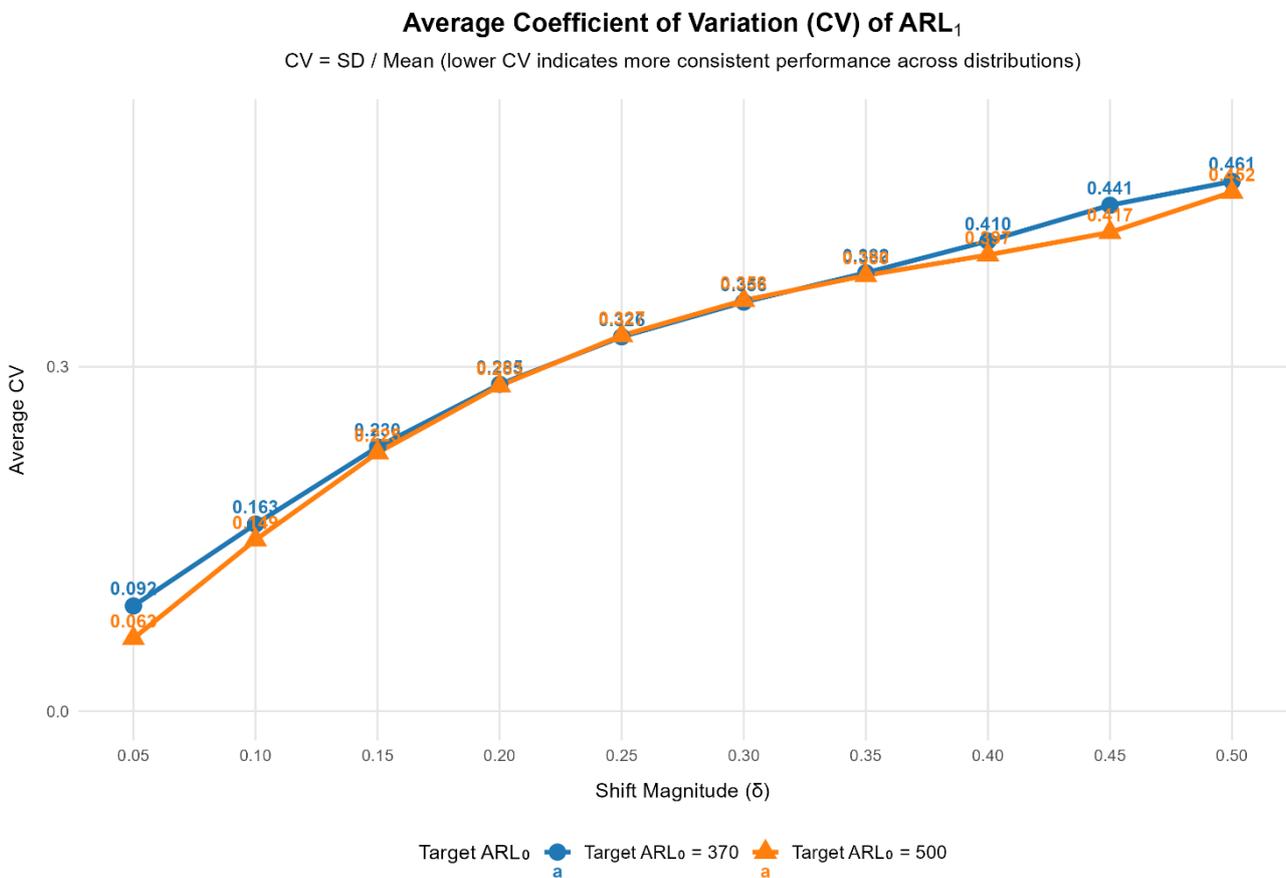

Figure 2: Average Coefficient of Variation (CV) of ARL$_1$ across the Normal, Laplace, Uniform, and Exponential distributions. CV = SD / Mean, where a lower CV indicates more consistent performance. Results are shown separately for targets of ARL$_0$ = 370 and ARL$_0$ = 500.)

Figure 2 shows that the CV analysis confirms high consistency. For the smallest detectable shift (δ = 0.05), the CV is exceptionally low at 0.065 and 0.090 for ARL$_0$ targets of 500 and 370, respectively. In statistical quality control, a CV below 0.10 is typically considered to indicate excellent consistency. While the CV increases with shift magnitude, this trend is a mathematical artifact of the ARL$_1$ metric. As the mean ARL$_1$ becomes very small (approaching 1 for large shifts), even minuscule absolute differences between distributions result in a larger relative CV, despite all distributions signalling an alarm in nearly the same, very short number of samples. Therefore, the rising CV for larger δ does not indicate deteriorating practical robustness but rather reflects the metric's scale when detection is swift and near-perfect across all conditions, as seen in Figure 3

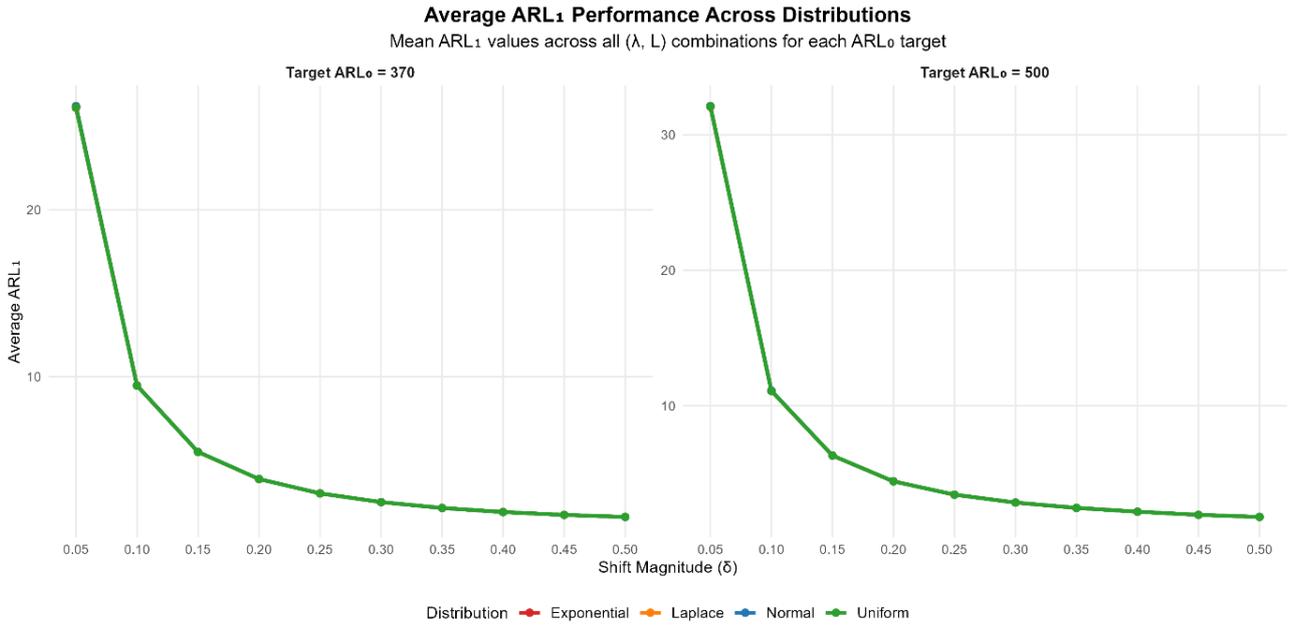

Figure 3: ARL$_1$ values across all ($\lambda$, L) combinations for the four tested distributions for different shifts. (a): Target ARL$_0$ = 370. (b): Target ARL$_0$ = 500.

This multi-faceted validation in Figure 3, which combines visual overlap, perfect agreement at key shifts, and low CV metrics, robustly confirms that the CSB-EWMA chart delivers highly consistent and predictable performance, regardless of the underlying data-generating mechanism.

*4.2 Practical Guidance on Parameter Selection*

For effective implementation of the CSB-EWMA chart, parameter selection should be guided by the expected shift magnitude and desired false-alarm rate. To detect small to moderate shifts ($\delta < 0.20$), choose $\lambda$ in the range $0.1 - 0.3$; for larger shifts ($\delta \geq 0.30$), values between $0.3 - 0.6$ balance sensitivity and robustness. A default of $\lambda = 0.2$ is recommended when shift size is unknown. The control limit multiplier $L$ should be selected inversely to $\lambda$ to maintain the target in-control ARL: for ARL$_0$ = 370, use $L \approx 1.4$ when $\lambda = 0.2$, and $L \approx 1.575$ when $\lambda = 0.5$. For stricter false-alarm control (ARL$_0$ = 500), increase $L$ by approximately $0.1 - 0.15$. Optimal $(\lambda, L)$ pairs are tabulated in Table 2.

The chart is robust to moderate variations in these parameters; sensitivity analysis shows that deviations of $\pm 0.05$ in $\lambda$ or $\pm 0.1$ in $L$ typically alter ARL$_0$ by less than 10%. For a balanced default configuration targeting ARL$_0 \approx 370$, we recommend $\lambda \approx 0.2$ and $L \approx 1.4$. For stricter control (ARL$_0 \approx 500$), use $\lambda \approx 0.15$ and $L \approx 1.55$. These settings perform well across diverse stream counts and distributions.

## DISCUSSION AND CONCLUSION

This study introduced the Cumulative Standardized Binomial EWMA (CSB-EWMA) chart for distribution-free monitoring of Multiple Stream Processes. By deriving the exact time-varying variance of the EWMA statistic for dichotomized data relative to the in-control median, the method provides adaptive control limits that are statistically valid from the start of monitoring, overcoming a key limitation of asymptotic approximations. Simulations confirm that the CSB-EWMA chart achieves rapid detection (ARL$_1$ of 3–6 samples for $\delta = 0.20$) while maintaining target in-control performance (ARL$_0 \approx 370$ or 500). It also demonstrates strong robustness across diverse distributions (Normal, Laplace, Uniform, Exponential), with low Coefficients of Variation (CV < 0.10) for out-of-control ARL.

The current work assumes independent streams and a known in-control median. Future research could extend the method to correlated streams, adapt it for Phase I use or estimated parameters, and evaluate performance under asymmetric shifts. Real-world case studies would further validate its practical utility. The CSB-EWMA chart effectively bridges the nonparametric MSP framework with the sensitivity of EWMA, offering practitioners a rigorous and responsive tool for early change detection in binomial multiple-stream processes

**Code**
Available in Gitub.

**DISCLOSURE STATEMENT**
The authors report there are no competing interests to declare.